\def\year{2019}\relax
\documentclass[letterpaper]{article} 
\usepackage{aaai19}  
\usepackage{times}  
\usepackage{helvet}  
\usepackage{courier}  
\usepackage{url}  
\usepackage{graphicx}  
\usepackage{booktabs}       
\usepackage{amsfonts}       
\usepackage{nicefrac}       
\usepackage{microtype}      
\usepackage{amsmath}
\usepackage{amsthm}
\newtheorem{definition}{Definition}
\usepackage{mathtools}
\usepackage{multirow}
\begin{document}
%
\title{Number Sequence Prediction Problems for Evaluating Computational Powers of Neural Networks}
\author{
  Hyoungwook Nam\\
  College of Liberal Studies\\
  Seoul National University\\
  Seoul, Korea \\
  \texttt{hwnam831@snu.ac.kr} \\
  \And
  Segwang Kim\\
  Department of Electrical and\\
  Computer Engineering\\
  Seoul National University\\
  Seoul, Korea \\
  \texttt{ksk5693@snu.ac.kr} \\
  \And
  Kyomin Jung\\
  Department of Electrical and\\
  Computer Engineering\\
  Seoul National University\\
  Seoul, Korea \\
  \texttt{kjung@snu.ac.kr} \\
}
\maketitle
\begin{abstract}
Inspired by number series tests to measure human intelligence, we suggest number sequence prediction tasks to assess neural network models' computational powers for solving algorithmic problems.
We define the complexity and difficulty of a number sequence prediction task with the structure of the smallest automaton that can generate the sequence.
We suggest two types of number sequence prediction problems: the number-level and the digit-level problems.
The number-level problems format sequences as 2-dimensional grids of digits and the digit-level problems provide a single digit input per a time step.
The complexity of a number-level sequence prediction can be defined with the depth of an equivalent combinatorial logic, and the complexity of a digit-level sequence prediction can be defined with an equivalent state automaton for the generation rule.
Experiments with number-level sequences suggest that CNN models are capable of learning the compound operations of sequence generation rules, but the depths of the compound operations are limited.
For the digit-level problems, simple GRU and LSTM models can solve some problems with the complexity of finite state automata.
Memory augmented models such as Stack-RNN, Attention, and Neural Turing Machines can solve the reverse-order task which has the complexity of simple pushdown automaton.
However, all of above cannot solve general Fibonacci, Arithmetic or Geometric sequence generation problems that represent the complexity of queue automata or Turing machines.
The results show that our number sequence prediction problems effectively evaluate machine learning models' computational capabilities.
\end{abstract}

\section{Introduction}
Well-defined machine learning tasks have been crucial for machine learning researches.
Major deep learning breakthroughs in the field of computer vision such as AlexNet \cite{alexnet}, VGGNet \cite{vgg} and ResNet \cite{he2016deep} could not be possible without Imagenet dataset and challenges \cite{imagenet}.
In the field of reinforcement learning, open-source platforms like MuJoCo \cite{mujoco} and Deepmind Lab \cite{deepmindlab} provide challenging environments for the studies.
However, it is hard to find machine learning task suite for algorithmic reasoning although reasoning has always been a significant subject for many machine learning studies.

It is theoretically proven that carefully designed neural network models can simulate any Turing machine \cite{siegelmann1995computational}.
Hence, there have been studies applying neural network models to solve algorithmic tasks such as learning context-sensitive languages \cite{gers2001lstm}, solving graph questions \cite{graves2016hybrid}, and composing low-level programs \cite{reednpi}.
Also, there have been attempts to train neural networks with simple numerical rules such as copy, addition or multiplication \cite{stackrnn,neuralgpu,gridlstm,neuralturing}.
However, it has been unclear whether the proposed models express computational powers equivalent to Turing machines in practice.
To provide a method to test the computational powers of neural network models, we propose a set of number sequence prediction problems designed to fit deep learning methods.

A number sequence prediction problem is a kind of intelligence test for machine learning models inspired by number series tests, which are conventional methods to evaluate non-verbal human intelligence \cite{nonverbal}.
A typical number series test gives a sequence of numbers with a certain rule and requires a person to infer the rule and fill in the blanks.
Similarly, a number sequence prediction problem requires a machine learning model to predict the following numbers from a given sequence.
The numbers are represented as a sequence of digit symbols; hence the model has to learn discrete transition rules between the symbols such as carry or borrow rules.

To be specific, we suggest two types of number sequence prediction problems: the number-level problems and the digit-level problems.
A number-level problem provides a two-dimensional grid of digits as an input where each row of the grid represents a multi-digit number.
The target would be a grid of the same format filled with the following numbers.
Solving a number-level problem is equivalent to constructing the combinatorial logic for the transition rules.
On the other hand, a digit-level task provides a single digit as an input per each time step.
A model needs to simulate a sequential state automaton to predict the outputs.
The type of the state machine required can vary from a finite state machine to a Turing machine, depending on the generation rule of the sequence.

The number sequence prediction problems are good machine learning tasks for several reasons.
First, typical deep learning models can easily fit into the problems.
Generative models for 2D images can be directly applied to solve the number-level problems, and recurrent language models can fit into the digit-level problems after minimal modifications.
Next, it is possible to define the complexity and difficulty of the problem.
Like Kolmogorov complexity \cite{chaitin1977algorithmic}, we can define the complexity of a problem with the structure of the minimal automaton needs to be simulated.
Finally, we can generate an arbitrarily large number of examples, which is hard for many machine learning tasks.

To empirically prove that the number sequence prediction problems can effectively evaluate the computational capabilities of machine learning models, we conduct experiments with typical deep learning methods.
We apply residual convolution neural network (CNN) \cite{he2016deep} models for the number-level problems, and recurrent neural network (RNN) models with GRU \cite{gru} or LSTM \cite{hochreiter1997long} cells to the digit-level problems.
We also augment RNN models with stack \cite{stackrnn}, external memory \cite{neuralturing} and attention \cite{bahdanau2014neural} which might help models solve more complex digit-level sequence prediction tasks.
One-dimensional CNN models can be applied to digit-level sequences, but it is not equivalent to solving digit-level problems because for the CNN models the data needs to be given at the same time in parallel, losing the sequential nature of the problems.
For each type of sequences, we measure the complexity of it by designing an automaton equivalent to the generation rule.
In the experiments of the number-level problems, sequences are generated by various linear homogeneous recurrence relations.
Since the digit transition rules of the relations can be implemented with combinatorial logic, we measure the complexity and the difficulty of a sequence from the width and the depth of the logic.
Experiments show that CNN models are capable of learning the compound operations of number-level sequence generation rules but limited to certain complexity.
Digit-level sequence prediction problems can be solved with state automata.
Therefore, we define the complexity of a problem with the computing power of the automaton and choose sequences with complexities of finite state automata, pushdown automata, and linear bound Turing machines.

The contributions of this work are as follows:
\begin{itemize}
\item We propose a set of number sequence prediction problems for evaluating a machine learning model's algorithmic computing power.

\item We define methods to measure complexities and difficulties of the problems based on the structure of automata to be simulated, which can predict the difficulty of training.

\item Number-level sequence prediction experiments show that CNN models can simulate deep combinatorial logics up to certain depth.

\item Digit-level sequence prediction tasks reveal that the computational powers of existing recurrent neural network models are limited to that of finite state automata or pushdown automata.

\end{itemize}

Overall, the set of our problems can be a well-defined method to verify whether a new machine learning architecture extends the computing power of previous models.
There are some possible directions to extend the computational capabilities of neural network models.
The first way is to apply training methods other than the typical methods we used in the experiments.
For instance, reinforcement learning methods can be applied to the algorithmic tasks \cite{zaremba2016learning}.
Next, non-backpropagation methods such as dynamic routing \cite{capsnet} might help neural network models learn more complex rules.
Our number sequence prediction tasks would provide a well-defined basis for those possible future works.

\section{Problem Definition}


\subsection{Number-level Sequence Prediction}

\begin{figure}[htbp]
  \centering
    \includegraphics*[width=0.65\linewidth]{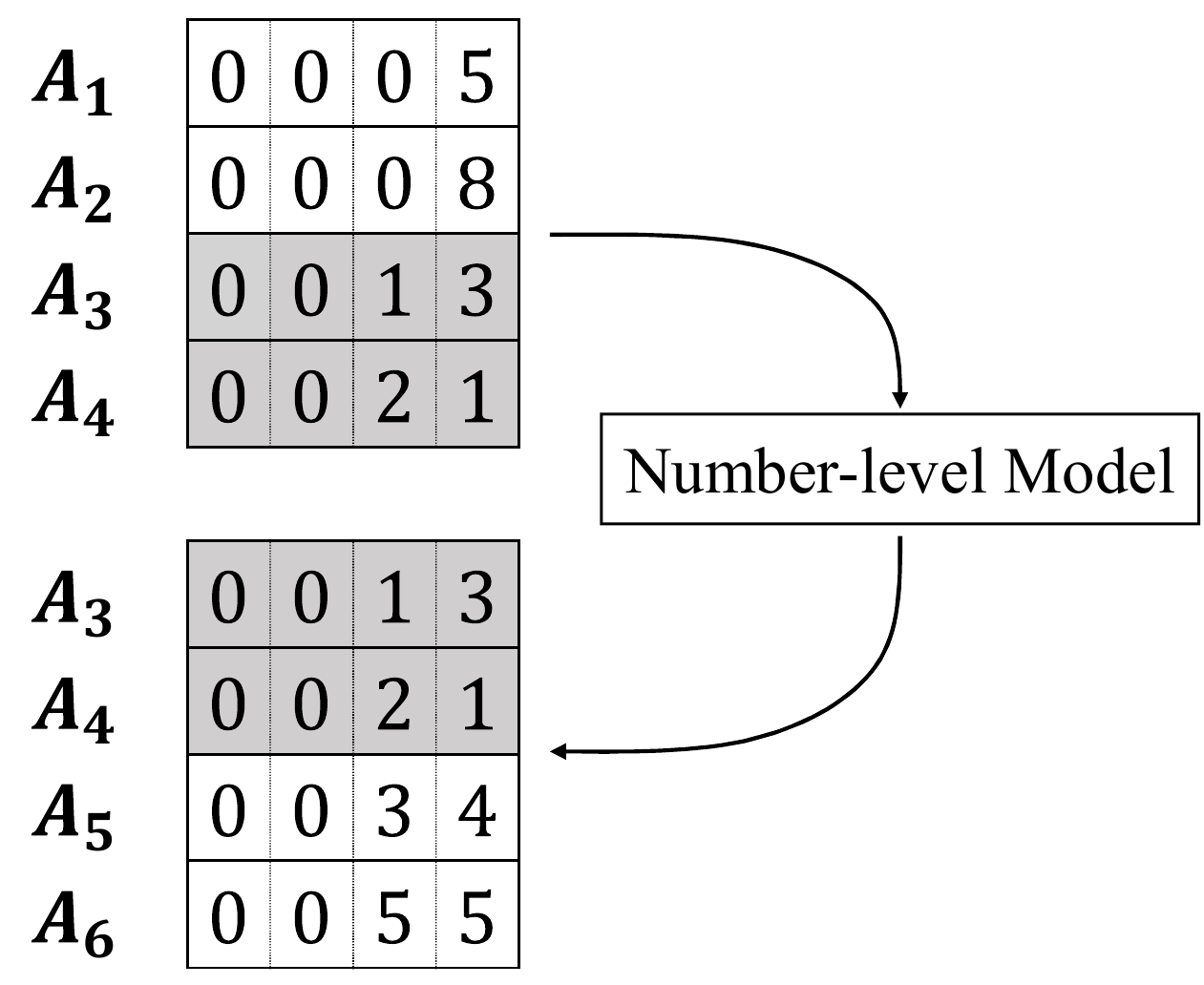}
\caption{
Input and target sequence examples of a number-level problem with the Fibonacci sequence. The number-level sequence example is with length $n=4$, shift $s=2$ and digit $l=4$. A number in a cell is represented by an one-hot vector.
\label{fig_numberproblem}
  }
\end{figure}

Figure~\ref{fig_numberproblem} illustrates a number-level sequence prediction problem.
The model is given with an input sequence $A_{1} \cdots A_{n}$ which is formatted as a two-dimensional grid with $n$ rows.
Each row corresponds to a term $A_{i}$ which is a multi-digit number of $l$ digits.
A digit cell is a one-hot vector where the number of channels is equal to the base $b$ of the digits.
The target data $A_{n+1} \cdots  A_{n+s}$ is a sequence of following numbers with the length shift $s$ with the same data layout.
In the experiments, we use sequence data of $n=8$, $l=8$ and $s=4$.
We denote this $\{0,1\}^{l\times b}$ binary one-hot row tensor representation of a natural number $A$ as $\langle A\rangle$.

We use various order-$k$ homogeneous linear recurrence of the form $A_n=c_1 A_{n-1}+\cdots+c_k A_{n-k}$ with constant integer coefficients $c_1, \dots, c_k$ to generate number sequences starting from randomly selected initial terms $A_1, \dots,  A_k$.
For instance, $k=2, \ c_1=1, \ c_2=1$ imply a general-Fibonacci sequence and $k=2, \ c_1=2, \ c_2=-1$ give an arithmetic progression.
Likewise, a progression with arithmetic sequence as its difference whose recurrence is $A_{n}-A_{n-1}=A_{n-1}-A_{n-2}+c$ can be re-written in $A_{n}=3A_{n-1}-3A_{n-2}+A_{n-3}$.
In the perspective of combinatorial logic, the generation rules of the sequences can be seen as $k$-ary operations of the binary tensors.
For example, the generation rule of arithmetic sequences can be represented with a binary operation of $(\langle A\rangle, \langle B\rangle) \mapsto \langle 2A-B\rangle$.
Since all inputs and outputs of an operation are binary, there exists a shortest disjunctive normal form (DNF) for the operation.
We first define a combinatorial width of an operation with its disjunctive normal form, i.e. sum of minterms\footnote{A logical AND of literals in which each variable appears exactly once in true or complemented form \cite{logicdesign} }.
 
\begin{definition}
If the smallest DNF of a function $f : \{0,1\}^n \rightarrow \{0,1\}^m$ has $\Theta(w)$\footnote{$w$ is a function of input and output dimensions} minterms, $\Theta(w)$ is called the \textbf{combinatorial width} of the function. If functions $f_1, \dots, f_k$ have corresponding widths of $\Theta(w_1), \dots, \Theta(w_k)$, the \textbf{compound width} of a composition $f_1 \circ \dots \circ f_k$ is defined as $\Theta(w_1+\dots+w_k)$.
\end{definition}

\begin{figure}
  \centering
  \includegraphics*[width=0.7\linewidth]{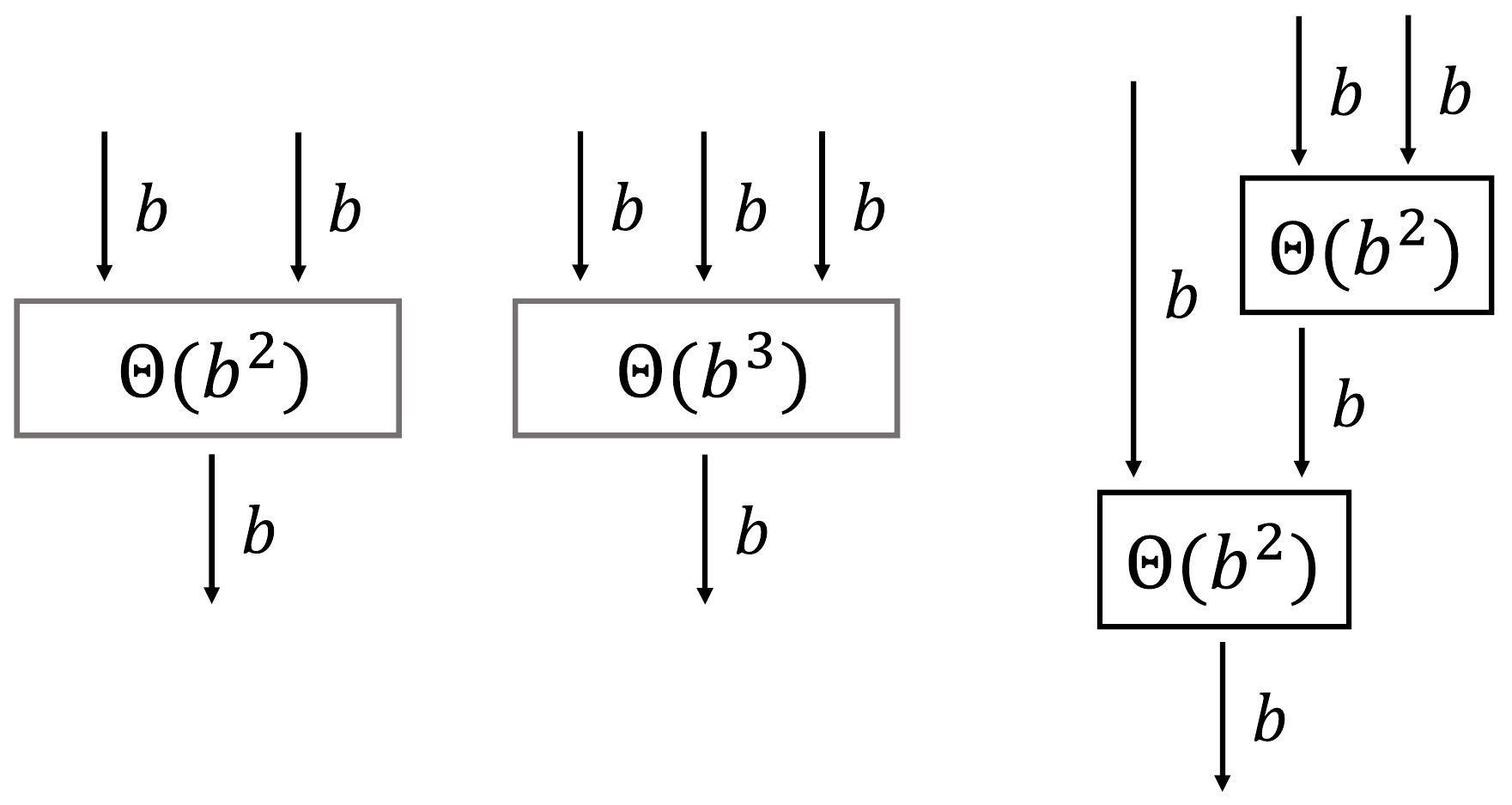}
  \caption{
Conceptual schema of a binary operation (left), a ternary operation (middle) and an equivalent composition of two binary operations (right). The formulas represent combinatorial widths.
\label{fig_logic}
  }
\end{figure}

The decimal digit addition, for example, requires at least $\Theta(10^2)$ products since it has to memorize the consequences of all possible digit pair inputs.
Therefore, the combinatorial width of a linear binary operation is $\Theta(b^2)$ where $b$ is the base of the digits.
Note that the compound width of a function is not unique.
Consider a logical circuit for the ternary operation of $(\langle A\rangle, \langle B\rangle, \langle C\rangle) \mapsto \langle 2A-B+C\rangle$.
As seen in Figure~\ref{fig_logic}, the operation can be implemented with a single function of combinatorial width $\Theta(b^3)$, or a compound of two binary operations resulting in the compound width of $\Theta(b^2)$ in at the cost of a deeper data path.
This depth of the path can define the complexity of the operation.

\begin{definition}
The \textbf{complexity} of a function $f : \{0,1\}^n \rightarrow \{0,1\}^m$ is the minimum number $n$ of functions which make the compound width of $f_1 \circ \dots \circ f_n = f$ the smallest. Such smallest compound width is called the \textbf{difficulty} of the function.
\end{definition}

For example, the length of a row $l$ is the complexity of the carry rule since the carry digit of the most significant digit sequentially depends on all other digits.
To eliminate the dependence on the dimensions, we ignore the carry or borrow rule while calculating a complexity.
Since a logical product can be approximated with a neuron with a nonlinear activation, the difficulty should correspond to the number of neurons in the network.
Also, since the complexity reflects the depth of a logical circuit, it should correspond to the number of layers in the network.
Note that it is possible to compromise the width for the depth as seen in Figure~\ref{fig_logic}.
We expect deep neural networks to learn narrow but deeper representations.

\subsection{Digit-level Sequence Prediction}

\begin{figure}[htbp]
  \centering
    \includegraphics*[width=0.8\linewidth]{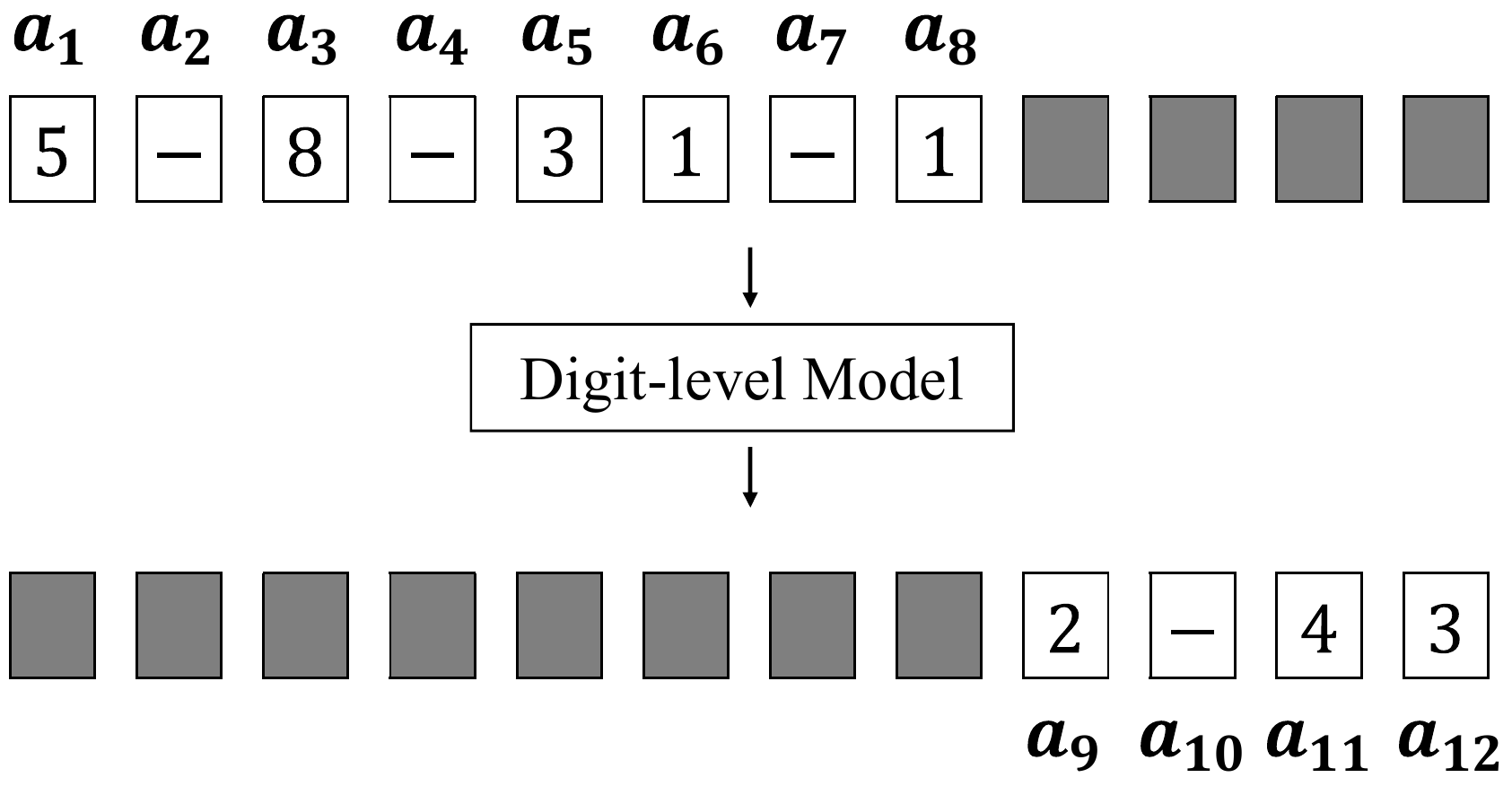}
\caption{
Input and target sequence examples of a digit-level problem with the Fibonacci sequence. The example is with $n=8$ and $s=4$. The order of the digits is little-endian (least significant digits first).
\label{fig_digitproblem}
  }
\end{figure}

Figure~\ref{fig_digitproblem} illustrates a digit-level sequence prediction problem.
The model is given with sequential inputs of $a_{1} \dots a_{n}$, each of which is an integer number corresponds to a character. 
With the base of $b$, the numbers $0 \dots b-1$ correspond to the digits.
The second last number $b$ is a blank, and the last number $b+1$ is a delimiter.
After $n$ inputs, we give delimiters as inputs for $s$ time steps.
The target sequence consists of $n$ delimiters followed by $a_{n+1} \dots a_{n+s}$.
Because digit calculations must start from the smallest digit, we order the digits in the little-endian order which is the reverse of the typical digit order.
In the experiments, we use sequences of $n=12$ and $s=12$.

The sequential nature of the data makes it more difficult to solve the problems.
Since the model has to retain information from the previous inputs, solving the problem is equivalent to modeling a sequential state automaton of the generation rule.
The computing power of a state automaton falls into one of the four categories: finite state machine, pushdown automaton, linear bounded automaton, and Turing machine.
All Turing machines are linearly bounded in the problems because the computation time is linearly bounded to the length of the sequence.
Therefore, three levels of state automata are possible in the digit-level sequence prediction problems.
We define the complexity of a sequence by the smallest state automaton required.

\begin{definition}
The \textbf{complexity} of a number sequence prediction problem is the complexity of a state automaton which can simulate the sequence generation rule with the smallest number of states. The \textbf{minimal grammar} of the sequences is the formal grammar can be recognized with the automaton.
\end{definition}

To illustrate, we can think about the most straightforward sequence of number counting.
If the numbers have at most $l$ digits of base $b$, the counter can be implemented with $\Theta(lb)$ shift registers which can be translated to the same number of non-deterministic finite state automaton.
Hence, the complexity of counting numbers is the complexity of finite state automata, and its minimal grammar is a regular grammar.
In the experiments, we use progressions with a fixed difference because they can be understood as generalized forms of number counting sequences.
Arithmetic, geometric and general-Fibonacci sequences can also be represented as digit-level sequences.
The most straightforward automata capable of generating them are queue automata, which share the same computational powers with Turing machines.
Since Turing machines must be linearly bounded in the digit-level problems, the minimal grammars of both arithmetic and general-Fibonacci sequences are context-sensitive grammars.

\begin{figure}[htbp]
  \centering
  \includegraphics*[width=0.9\linewidth]{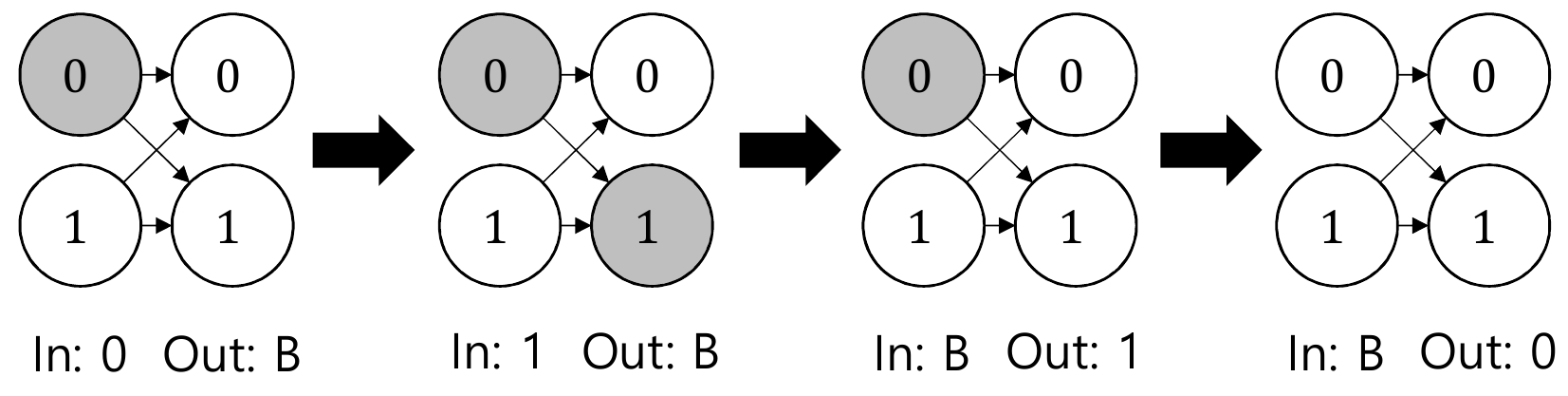}
  \caption{
Nondeterministic finite state automaton that can solve reverse-order task with $n=2$ and $b=2$. Automata for fixed difference arithmetic sequence can be built in similar manners.
\label{fig_fsm}
  }
\end{figure}

Between regular and context-sensitive languages, there are context-free languages which require pushdown automata.
Palindromes are proper examples of context-free languages which cannot be expressed by lesser languages.
Therefore, we add the experiment of a reverse-order task where the target sequence is the reverse of the input sequence.
The input data consists of $n$ random digits followed by $n$ delimiters, and
the target data is $n$ delimiters followed by $n$ digits, which is the reverse of the input sequence.
If $n$ is limited, it is possible to solve the reverse-order task with a finite state automaton as seen in Figure~\ref{fig_fsm}.
Therefore, we train the models with $n=1 \dots 12$ and validate with $n=16$ to force the complexity of the problem equivalent to a pushdown automaton.

\section{Method}

\subsection{Model Architecture}
\begin{figure}[htbp]
  \centering
  \includegraphics*[width=0.8\linewidth]{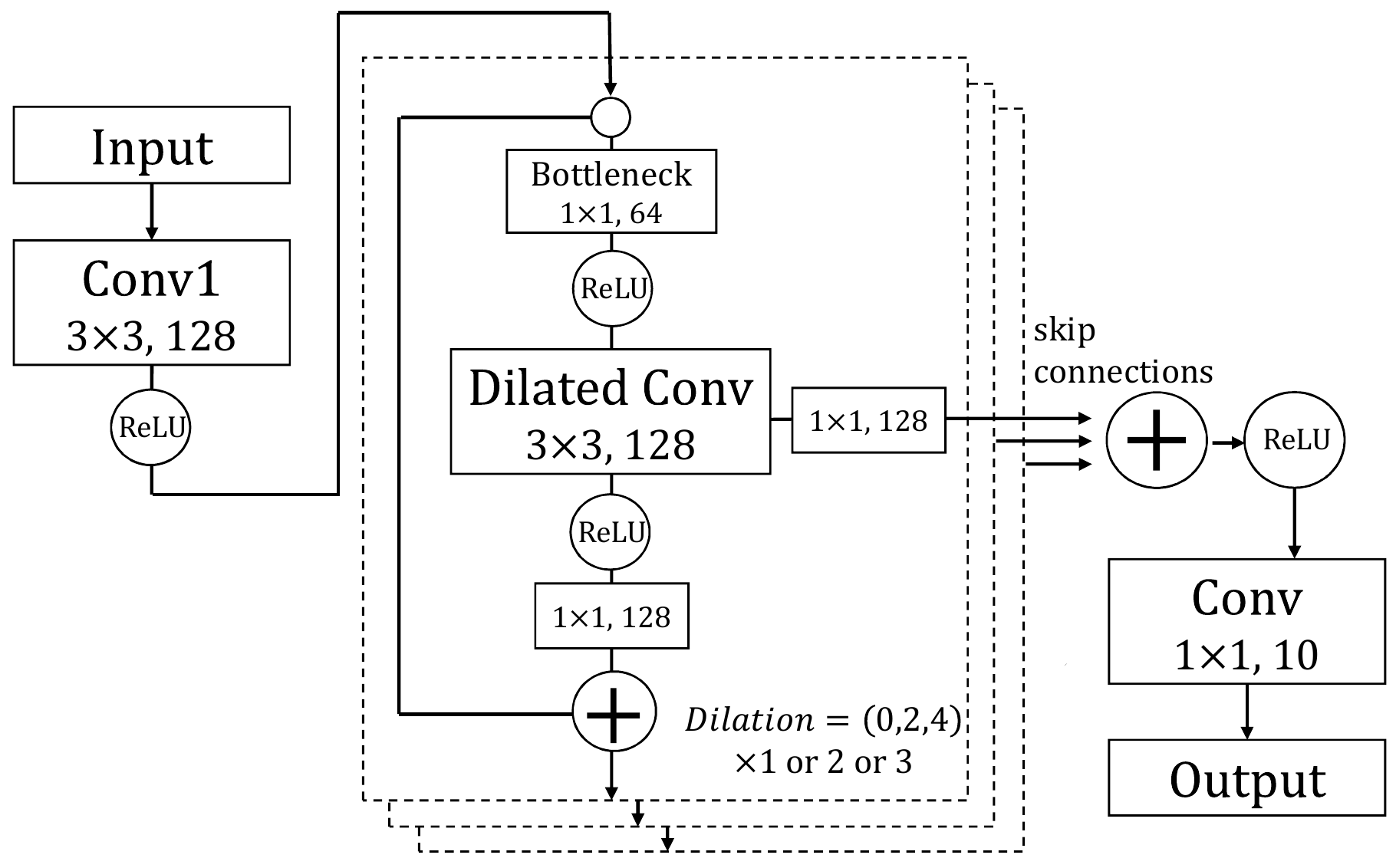}
  \caption{
Schematic of number-level CNN models. The number of neurons in the convolution layers can be one of ${64, 128, 192}$. The residual blocks can be repeated once, twice or thrice, making 12, 21 or 30-layer CNN model.
\label{fig_cnnmodel}
  }
\end{figure}

The number-level sequence prediction models described on Figure~\ref{fig_cnnmodel} are based on WaveNet model \cite{wavenet} which is also a generative model for sequential data.
Since the data layout of number-level sequences is two-dimensional, we use 3$\times$3 convolution kernels with dilation on the second dimension of the kernels where large receptive fields are necessary for the carry rules.
Unlike WaveNet, we use ReLU activation because we empirically observe that the gate activation slows down the training speed but shows no improvement on the accuracy.
The number of neurons per convolution layer can be 64, 128 or 192, which can correspond to the difficulty of a problem.
Inspired by the bottleneck architectures of residual CNN \cite{he2016deep}, the first layer of each residual block has half the number of neurons.
By stacking more residual blocks to the model, we can change the depth of the model.
The base 12-layer model has three residual blocks of dilation (0,2,4), and they can be repeated to make 21-layer and 30-layer models.
BatchNorm \cite{batchnorm} and Dropout \cite{dropout} methods are applied to all residual blocks.

\begin{figure}[htbp]
  \centering
  \includegraphics*[width=0.9\linewidth]{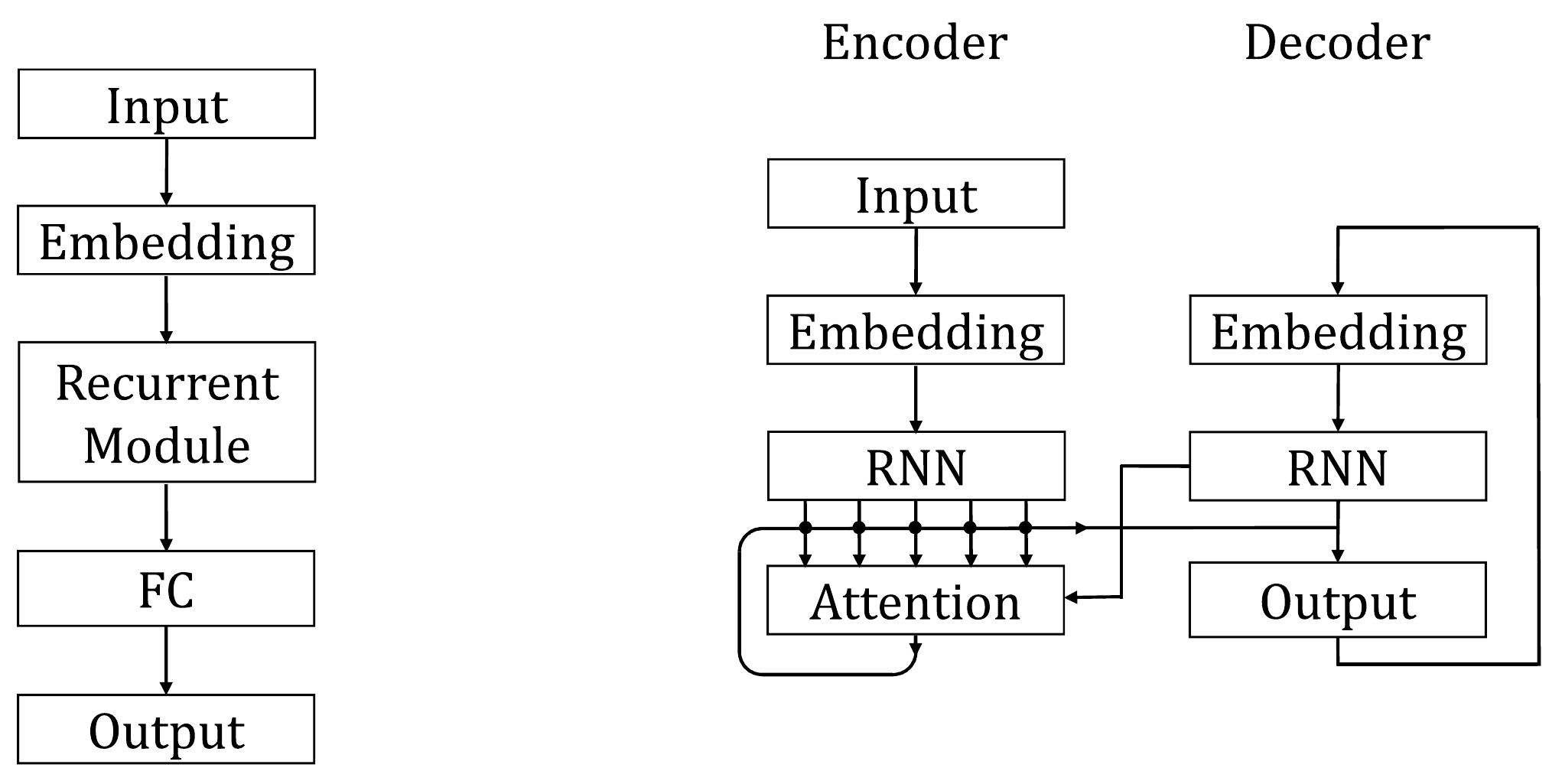}
  \caption{
Schematics of digit-level neural network models. A recurrent module in a digit-level model can be either LSTM, GRU, Stack-RNN or Neural Turing Machine. Unlike other digit-level models, an attention model must follow the encoder-decoder structure which is illustrated on the right side.
\label{fig_rnnmodel}
  }
\end{figure}

The digit-level sequence prediction models on the left side of Figure~\ref{fig_rnnmodel} are based on the simple character-level RNN language model \cite{char-rnn} with minimal modifications.
LSTM \cite{hochreiter1997long}, GRU \cite{gru}, Stack-RNN \cite{stackrnn} and Neural Turing Machine (NTM) \cite{neuralturing} are used for the recurrent modules in the middle.
A Stack-RNN module uses a number of stacks equal to the base $b$, and an NTM module uses 4 read and write heads.
A digit-level model with attention \cite{bahdanau2014neural} follows the encoder-decoder architecture on the right side of  Figure~\ref{fig_rnnmodel}.
The first half of an input sequence and the second half of a target sequence begun with the delimiter ($\langle$Go$\rangle$ symbol) are fed into the encoder and the decoder.
We use both unidirectional and bidirectional LSTM modules for the models with attention.
We set the number of neurons in all hidden layers to 128.

\begin{figure*}[tbp]
  \centering
  \includegraphics*[width=\linewidth]{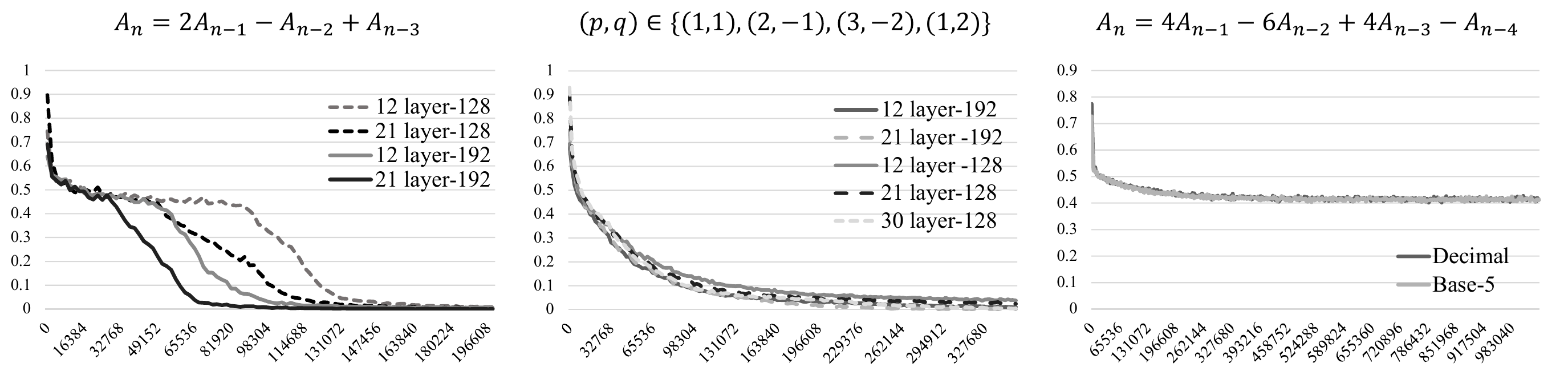}
  
  \caption{
Validation error curves of the deep and wide models on the sequences generated by a ternary relation (left), the mixture of the four types of binary relations (middle) and the quaternary relation with different base digits (right). Y-axes are validation error rates, and the X-axes are training example counts. The value following the number of layers denotes the number of neurons in a convolution layer. The third experiment uses a 30-layer model with 128 neurons per layer.
\label{fig_mixlv3}
  }
\end{figure*}

\subsection{Training and Validation Method}

We follow the end-to-end training fashion.
Thus the models have to learn the logical rules without any domain-specific prior knowledge.
A batch of size 32 is randomly generated for each iteration by choosing the initial numbers and applying the generation rules.
The space of all possible training sequences should be large enough to avoid overfitting. 
We evaluate the validation prediction error rate with a pre-defined validation dataset after every 32 iterations.
We define a prediction error rate as a ratio of wrong predictions to the total predictions.
The total predictions are counted as $l\times s = 32$ in number-level sequences and $s = 12$ in digit-level sequences.
A prediction is determined by the digit channel with the maximum output value.
Both number-level and digit-level models are trained to minimize the cross-entropy loss function.
The validation dataset is also randomly generated from the space outside of the training data space.
For example, we choose the first two terms of number-level arithmetic sequences from the range of $(0,20000)$ for the training dataset, but we choose them from the range of $(20000, 30000)$ for the validation dataset.

\section{Experiment}

\begin{figure}[htbp]
  \centering
  \includegraphics*[width=0.9\linewidth]{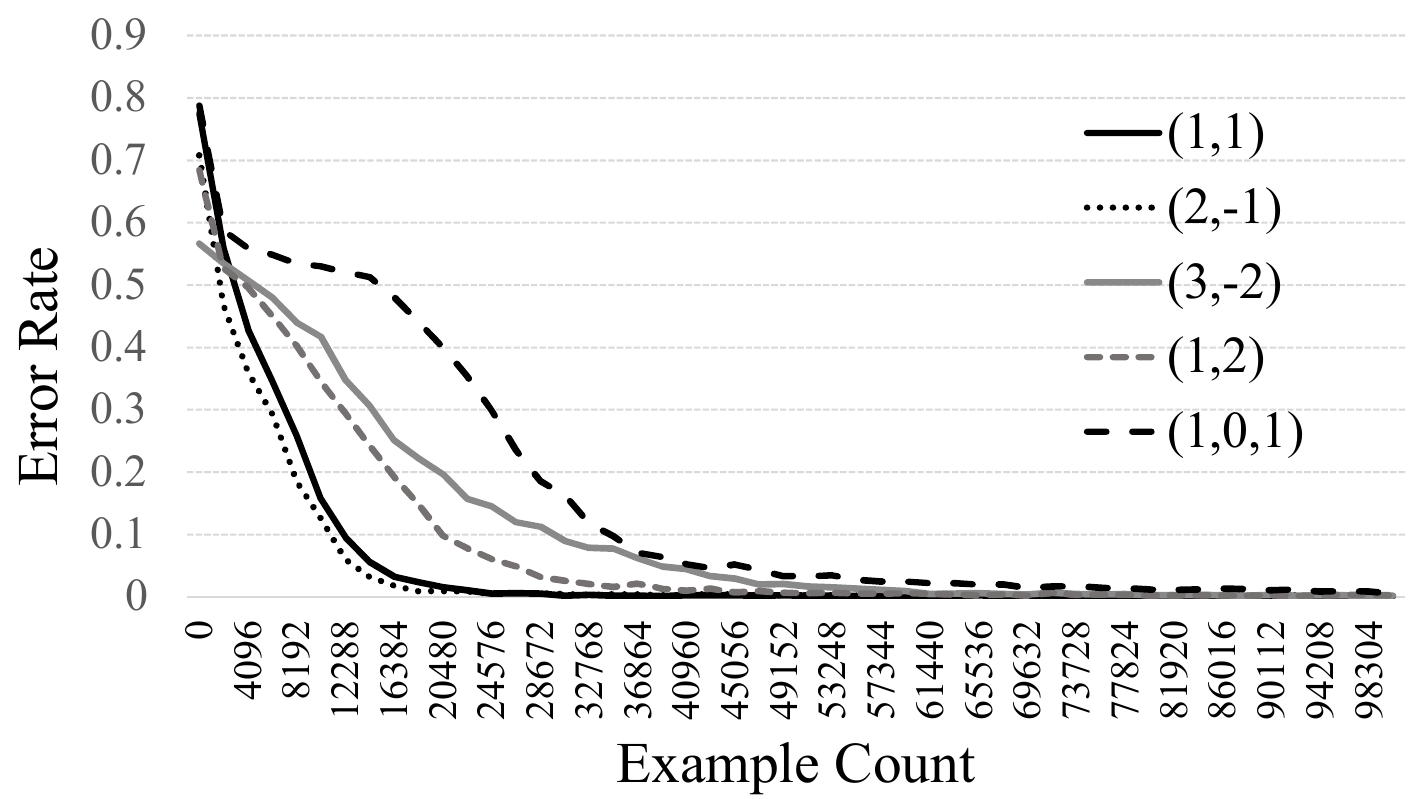}
  \caption{
The learning curves of the 12-layer number-level model with 64 neurons on the five types of the basic sequences. $(p,q)$ denotes the coefficients of binary operations $(A,B) \mapsto pA+qB$. $(1,0,1)$ denotes the relation of $A_n=A_{n-1}+A_{n-3}$.
\label{fig_lv2}
  }
\end{figure}

\subsection{Number-level Sequence Prediction Experiment}

\paragraph{Setup}
The objective of the experiments is to verify that complexity and difficulty of a number-level problem correspond to the depth and the parameter size of a CNN model.
Total eight types of sequences are used in this part of the experiments.
First four types of the sequences have recurrence relations in the form of $A_n=p A_{n-1}+q A_{n-2}$ where $(p,q)\in \{ (1,1),(2,-1),(3,-2),(1,2) \}$.
These four sequences represent binary operations with the complexity of one.
The fifth type of sequences has a relation of $A_n=A_{n-1}+A_{n-3}$ which represents a binary operation with the complexity of two because the model has to see through at least two layers to catch the relation between $A_{n-1}$ and $A_{n-3}$ with $3\times3$ convolution kernels.
The sixth type of the sequences is a mixture of the first four types of the sequences.
This is equivalent to building a ternary combinatorial logic with four times more width.
For comparison, the seventh type of sequences is generated by a recurrence relation of $A_n=2A_{n-1}-A_{n-2}+A_{n-3}$ which can be a compound of two binary operations.
The last type sequence is a progression with a relation of $A_n=4A_{n-1}-6A_{n-2}+4A_{n-3}-A_{n-4}$ whose general term can be calculated with a fourth-order polynomial.
For the training data, the first $k$ terms\footnote{$k$ is the order of a recurrence relation} of the sequences are chosen from $(0,20000)$, while they are chosen from $(20000,30000)$ in the validation dataset.
We compare the learning curve patterns over various model configurations.

\begin{figure}[htbp]
  \centering
  \includegraphics*[width=0.7\linewidth]{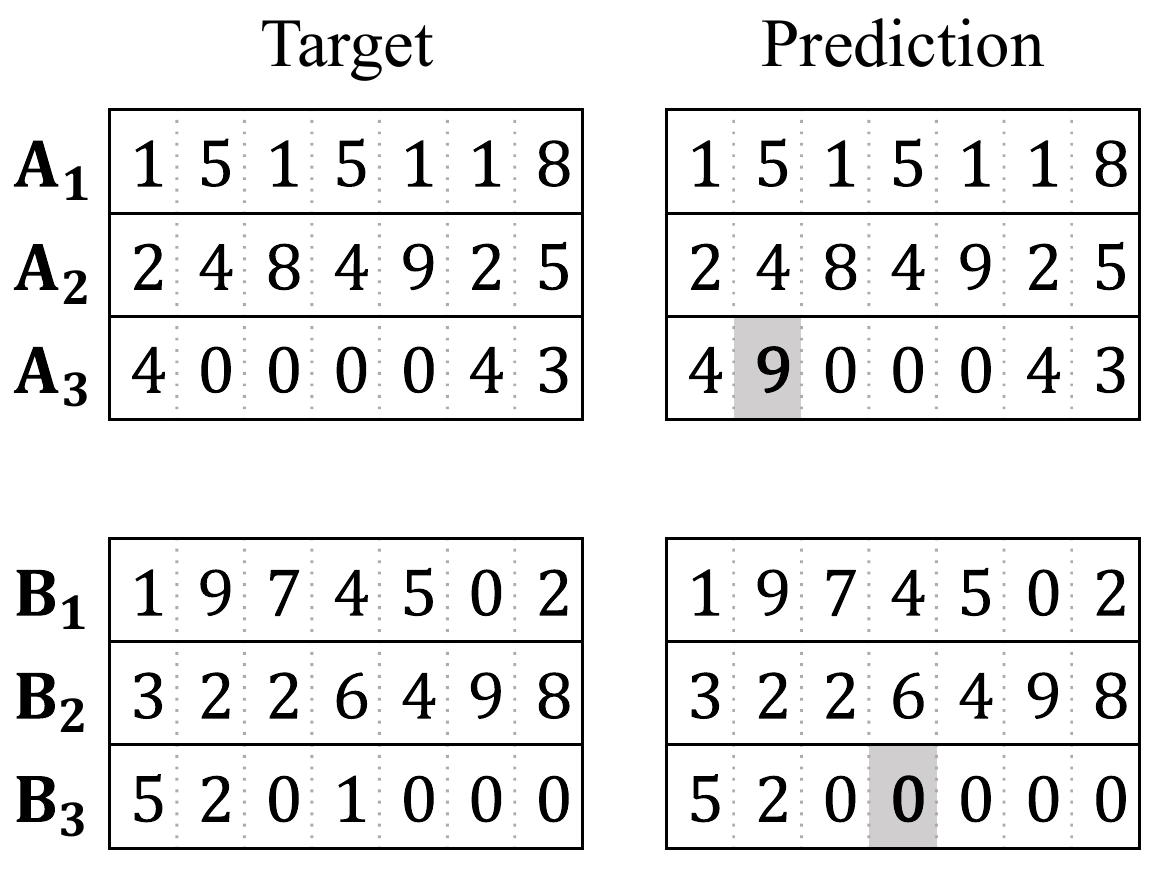}
  \caption{
The error examples from number-level model trained with general-Fibonacci sequences. Shaded cells show the locations of the errors. The numbers are shown in little-endian order.
\label{fig_lv2err}
  }
\end{figure}

\paragraph{Result}
Figure~\ref{fig_lv2} and Figure~\ref{fig_lv2err} show the validation error curves and the error examples of a CNN number-level model during the training on the five types of sequences generated by the binary operations.
Although the numbers of possible sequences exceed a hundred million, the model can achieve error rates near zero in less than a hundred thousand examples.
Since the validation data comes from the outside of the training data space, we can conclude that the model can learn the exact logic rules for the operations.
The error examples show that it is hard to catch long-term carry rules, which is expected because the carry rules have complexities equal to $l=8$.
Deploying deeper models reduce the errors from those long-term carry rules, occasionally achieving a zero prediction error.
The fifth sequence of rule $A_{n-1}+A_{n-3}$ shows a different learning curve pattern since $3\times3$ convolution kernels force the model to simulate the logic with the complexity of two.
The results show that complexities of number-level sequence prediction problems can effectively predict the hardness of learning.

Figure~\ref{fig_mixlv3} compares the learning curves of the models with various configurations and sequence data.
The models successfully learn the rules from both the mixed set of primary sequences and the sequences generated by a ternary relation.
However, the patterns of the learning curves are different.
With the mixed set of primary sequences, the learning curves of the models show uniform convexity without a saddle point.
Also, there is no clear advantage of using deeper models.
However, the learning curves with the sequences of a compound rule have saddle points, where we suspect the models find breakthroughs.
Moreover, we can observe the advantages of using deeper models.
Therefore, it can be concluded that deep learning models tend to learn complex but less difficult combinatorial logic, rather than the equivalent shallow but wide representations.
Meanwhile, the last learning curves show that the CNN model finds it hard to learn the logic with the complexities more than three.
The quaternary operator with base 5 has a smaller combinatorial width than a decimal ternary operator, but the model cannot learn the rule of the former.

\begin{table*}[tbp]
\centering
\def\arraystretch{1.5}
\begin{tabular}{|l|rrrr|}
\hline
Tasks     & Reverse-order (training) & Geometric & Arithmetic & Fibonacci \\ \hline
LSTM      & 28.4\% (1.2\%)	& 79.4\%    & 77.1\%     & 80.5\%    \\ 
GRU       & 51.9\% (0.9\%)	& 69.0\%    & 77.1\%     & 79.3\%    \\ 
Attention(unidirectional) & 42.0\% (8.8\%)  & 62.8\%    & 77.0\%     & 69.3\%    \\ 
Attention(bidirectional) & 0.0\% (0.0\%)  & 51.0\%    & 72.9\%     & 60.9\%    \\ 
Stack-RNN & \textbf{0.0\%} (0.0\%)     & 64.1\%    & 63.8\%     & 69.4\%    \\ 
NTM       & \textbf{0.0\%} (0.0\%)     & 57.1\%    & 65.7\%     & 68.1\%    \\ \hline
\end{tabular}
\caption{Test error rates of the digit-level sequence prediction experiment. Identical training methods are applied to the models except the attention model. Parenthesized numbers in the reverse-order task column are training error rates with $n=1\dots12$.}
\label{table_err}
\end{table*}

\begin{figure}[htp]
  \centering
  \includegraphics[width=0.9\linewidth]{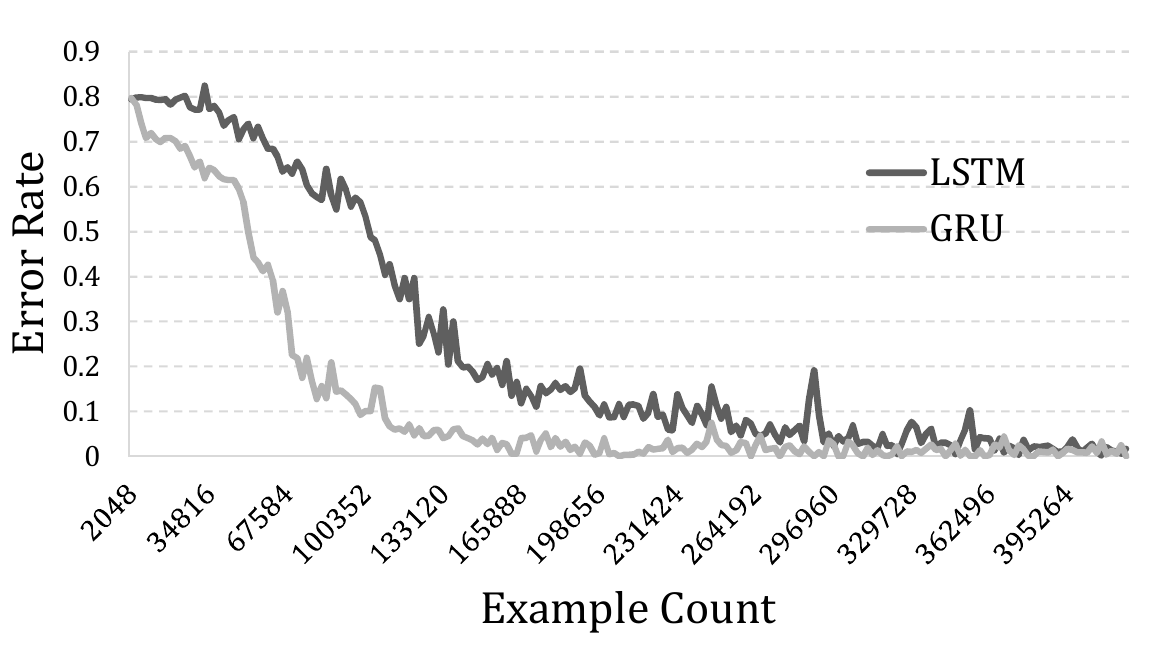}
  \caption{
Validation error curves of LSTM and GRU digit-level sequence prediction models on the arithmetic sequences with fixed difference of 17.
\label{fig_count}
  }
\end{figure}

\begin{figure}[htp]
  \centering
  \includegraphics[width=0.9\linewidth]{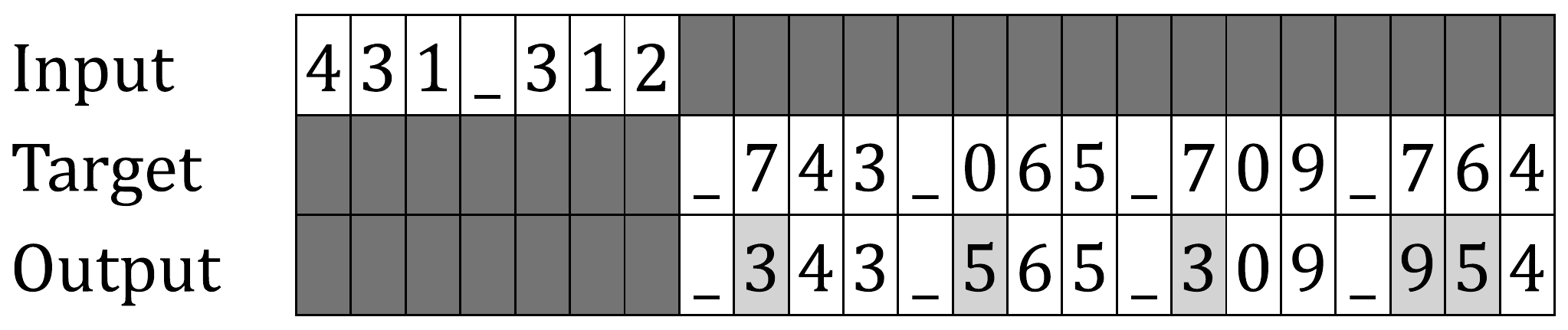}
  \caption{
Error examples from the digit-level LSTM model trained with general-Fibonacci sequences. The numbers are shown in little-endian order. Shaded cells show locations of the errors.
\label{fig_digitfib}
  }
\end{figure}

\subsection{Digit-level Sequence Prediction Experiment}

\paragraph{Setup}
The purpose of the digit-level sequence prediction experiments is to find complexity limits of the models.
The first type of the sequences is a progression with a fixed difference, which can be understood as a variation of number counting sequences.
We use the difference of 17 to observe the carry rules more often.
The first term of a training data is chosen from the range of $(0,9000)$, and that of a validation data is chosen from $(9000,9900)$. 
In the second experiment, we use arithmetic sequences or general-Fibonacci sequences.
The first two terms are chosen from the range of $(0,4000)$ during the training and $(4000,6000)$ for validations.
Since it is impractical to build finite state automata for all cases, the model must simulate queue automata to solve the problems.
The third experiment uses rounded geometric sequences with the relation $A_{n+1} = \lfloor1.3A_n\rfloor$ where  
the first terms are randomly chosen from $(0,4000)$ during the training and $(4000,6000)$ for validations.
The task also requires a smaller queue automaton since it has to remember only one previous number at a time.
The last experiment tests the models with the reverse-order task, which has the complexity of a pushdown automaton.
Since a reverse-order problem of fixed length can be solved by a finite automaton, we train the models with $n\in\{1 \dots 12\}$ and validate the models with $n=16$ to force the models to learn a pushdown automaton.

\paragraph{Result}
Figure~\ref{fig_count} shows that GRU and LSTM based models are capable of simulating finite state automata. 
Although the GRU model shows better performance than the LSTM model, both are not able to solve the problems that require queue or pushdown automata as seen in Table~\ref{table_err}.
Training error rates of GRU and LSTM models on reverse-order task converges around 0.01 suggesting that the models are capable of simulating finite state automata for generating palindromes with a limited length.
The error examples from the general-Fibonacci sequence prediction task in Figure~\ref{fig_digitfib} show the strategies of the models.
The models remember relationships between the most significant digits, while relationships between the least significant digits are more critical for the digit computations.
We can conclude that the computational powers of typical RNN models are limited to those of finite state automata if they are trained with typical training methods.
Encoder-decoder model with attention, Stack-RNN and NTM models are capable of solving reverse-order task, but they are no better than typical RNN models in problems that require queue automata.
The model with attention doesn't show significant differences if the model uses unidirectional LSTM.
Using bidirectional LSTM seems to be crucial for simulating pushdown automata in the models with attention.

\section{Conclusion}

We introduced effective machine learning problems of number sequence prediction which can evaluate a machine learning model's capability of solving algorithmic tasks.
We also introduced the methods to define the complexities and difficulties of the number sequence prediction problems.
The structure of a combinatorial logic measures the complexity and the difficulty of a number-level sequence prediction problem.
Experiments with the CNN models showed that they are effective ways to predict the hardness of learning, and they correspond to structures of neural network models.
The complexity of a digit-level sequence prediction task could be defined as the complexity of a minimal state automaton which can solve it.
The experimental results showed that the computational powers of typical RNN models are limited to those of finite automata.
While models augmented with external memory could solve the problems that require pushdown automata, none of the models were capable of simulating queue automata which are equivalent to Turing machines.
To sum up, our number sequence prediction tasks were proven to be effective and well-defined for testing neural network models' computational capabilities.

There are a few possible ways we suggest to proceed with the problems.
The first way is to propose and test a new network architecture to solve the tasks could not be solved in this study.
If a neural network model can solve digit-level arithmetic and geometric sequence prediction tasks, we can say that the model extended the computational capabilities of neural networks.
Another way is applying training methods other than typical methods we used in the experiments.
Sequence-to-sequence training methods for the digit-level prediction problems limit the capability of models to linear bound automata since the computation time is linearly bounded to the length of a sequence.
The training methods that decouple computation time and the number of outputs might expand the capability of the neural network models.
Finally, non-backpropagation methods such as dynamic routing \cite{capsnet} might be able to expand the computing power of neural network models.
Our number sequence prediction tasks would provide a well-defined basis for those possible future works.
\bibliographystyle{aaai}
\bibliography{aaai2019}
\end{document}